\documentclass[nohyperref]{article}

\usepackage{amsmath,amsfonts,bm}

\def\eqref#1{equation~\ref{#1}}

\def\1{\bm{1}}

\DeclareMathAlphabet{\mathsfit}{\encodingdefault}{\sfdefault}{m}{sl}
\SetMathAlphabet{\mathsfit}{bold}{\encodingdefault}{\sfdefault}{bx}{n}

\usepackage[accepted]{icml2022}
\usepackage{graphics}
\usepackage{hyperref}
\usepackage{url}
\usepackage{tabularx}
\usepackage{booktabs}
\usepackage{multirow}
\usepackage{xcolor}
\usepackage{soul}
\usepackage{color}
\usepackage{boxedminipage}
\usepackage{diagbox}
\usepackage{makecell}
\usepackage{microtype}
\usepackage{inconsolata}
\usepackage{enumitem}
\usepackage{graphicx}
\usepackage{booktabs}
\usepackage{xspace}
\usepackage{subcaption}
\usepackage{amsmath}

\usepackage{float}
\usepackage{tcolorbox}
\usepackage{wrapfig,booktabs}
\usepackage{tikz}
\usepackage{pgfplots}

\newcommand{\AD}{$\textsc{AdaptiveDecoder}$}
\newcommand{\ADtok}{$\textsc{AdaptiveDecoder}_{tok}$}
\newcommand{\ADseq}{$\textsc{AdaptiveDecoder}_{seq}$}
\newcommand{\smallADseq}{$\textsc{AD}_{seq}$}

\newcommand{\method}{LPO}
\newcommand{\methodmedium}{Latent Preference Optimization}
\newcommand{\methodlong}{Latent Preference Optimization}
\renewcommand{\vec}[1]{\boldsymbol{#1}}

\definecolor{my_purple}{HTML}{5617e8}

\icmltitlerunning{Adaptive Decoding via \methodmedium{}}

\begin{document}

\twocolumn[
\icmltitle{Adaptive Decoding via \methodmedium{}}

\icmlsetsymbol{equal}{*}

\begin{icmlauthorlist}
\icmlauthor{Shehzaad Dhuliawala}{meta,eth}
\icmlauthor{Ilia Kulikov}{meta}
\icmlauthor{Ping Yu}{meta}
\icmlauthor{Asli Celikyilmaz}{meta}
\\
\icmlauthor{Jason Weston}{meta}
\icmlauthor{Sainbayar Sukhbaatar}{meta}
\icmlauthor{Jack Lanchantin}{meta}
\end{icmlauthorlist}

\icmlaffiliation{meta}{FAIR at Meta.}
\icmlaffiliation{eth}{ETH Zürich}

\icmlcorrespondingauthor{Jack Lanchantin}{jacklanchantin@meta.com}

\icmlkeywords{Machine Learning, ICML}

\vskip 0.3in
]

\printAffiliationsAndNotice{} %

\begin{abstract}
During language model decoding, it is known that using higher temperature sampling gives more creative responses, while lower temperatures are more factually accurate.
However, such models are commonly applied to general instruction following, which involves both creative and fact-seeking tasks, using a single fixed temperature across all examples and tokens. In this work, we introduce {\em Adaptive Decoding}, a layer added to the model to select the sampling temperature dynamically at inference time, at either the token or example level, in order to optimize performance. To learn its parameters we introduce {\em \methodlong} (\method{}) a general approach  to train discrete latent variables such as choices of temperature. Our method outperforms all fixed decoding temperatures across a range of tasks that require different temperatures, 
including UltraFeedback, Creative Story Writing, and GSM8K. %
\end{abstract}

\section{Introduction}

\begin{figure*}[t!] 
\centering  
\scalebox{1.0}{ %
    \includegraphics[width=\textwidth]{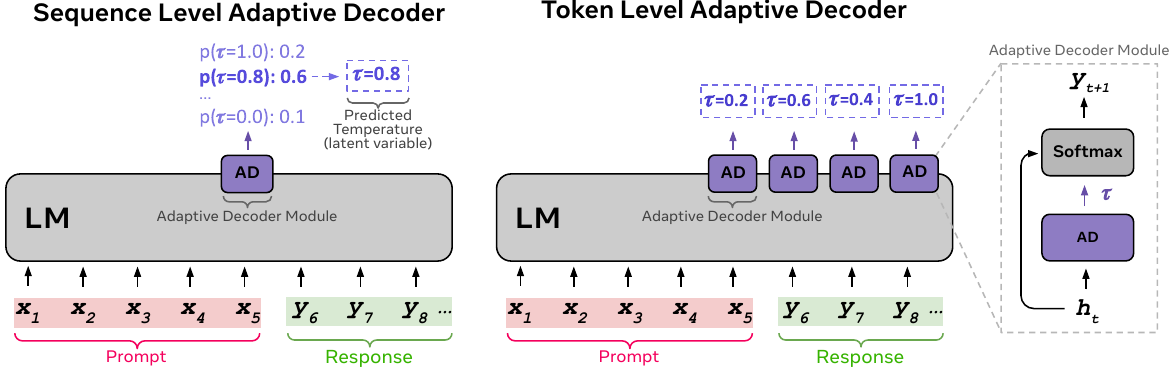}  
}
\caption{{\bf The \AD{}}. This learned module is added to the standard transformer in order to select decoding hyperparameters. It consists of a new decoder head attached to the last hidden state which assigns probabilities to different hyperparameter choices per token (right) or sequence (left), and the highest probability choice is selected in each case. This allows the LLM to select low temperatures for tokens requiring factual consistency, and higher temperatures for tasks requiring creativity and diversity. For the token level adaptive decoder,  a different temperature can be selected for different parts of the response given a single instruction.}
\label{fig:AdaptiveDecoder}  
\end{figure*}

 Large language models (LLMs) are powerful generalist models that can be used on a wide variety of tasks, ranging from fine-grained reasoning to open-ended creative writing \citep{openai2023gpt4,dubey2024llama}.
Yet, early works showed that after training, the decoding method still has a large effect on performance across these tasks, leading to the proposal of various temperature sampling techniques \citep{holtzman2019curious,welleck2019neural,fan2018hierarchical}.
In current LLMs, \textit{temperature} \citep{ackley1985learning} is a key post-training parameter for generation. Temperature is used to scale the next token probabilities to be either more uniform or more sharp. Lower temperature leads to less creative, more factual generations, and higher temperature leads to more creative and original generations.
Certain tasks, such as math problems or factual question answering, require the model to optimize accuracy of a single correct solution, and benefit from a low temperature, or greedy decoding \citep{shi2024thorough}. Others, like story generation, benefit from diverse and creative outputs, and a high decoding temperature. Intuitively, a complex task involving a number of these requirements might thus benefit from an adaptive temperature for different parts of its solution.

Existing LLM evaluation pipelines often rely on a fixed choice of temperature which is therefore suboptimal on some tasks, or else manual tuning is used to control the level of diversity in LLMs, which can be time-consuming, task-specific, and limited in its ability to adapt to changing requirements and prompts. 
To overcome this limitation, we introduce {\em Adaptive Decoding}, which consists of a new learnable layer, as well as a novel method to train it. 
The new learnable neural layer, which we call the \AD{}, is added to the final layers of the transformer architecture, enabling the LLM to dynamically adjust its  output diversity based on context (i.e, the task at hand). Specifically, the \AD{} allows the model to select an ideal temperature for decoding  the next token by adding a new decoder head attached to the last hidden state. We can either apply this at the example (sequence) level where a single temperature is predicted for all generated tokens, or the token level where a new temperature is predicted for each generated token.

Training the
\AD{} layer
requires discrete optimization over latent variables (i.e., the choice of temperature). In order to make this feasible, we introduce a general method for such training, called \methodlong{} (\method{}).
\method{} involves sampling multiple responses from the model, where the \AD{} layer will select temperatures (latent variables) that will affect the final token choices. Those responses are then  evaluated by a reward model in order to build chosen and rejected preference pair examples. Given these pairs, we use the \method{} loss to learn the optimal parameters of the \AD{} layer for selecting temperatures during decoding. 
Our approach thus learns the hyperparameters of generating text across diverse tasks, allowing the model to balance exploration and exploitation in a task-aware manner.

To validate our method, we experiment on a diverse set of tasks, ranging from math reasoning to creative writing and general instruction following.
We show that the decoder learns to select low temperatures for reasoning tasks like math, higher temperatures for creative writing, and somewhere in between for general prompts.
We find that when the training data includes all types of tasks, the model adaptively adjusts the temperature to the ideal value for each task by conditioning output token temperature choices on  the input context. This enables the \AD{} to be incorporated as part of the standard post-training pipeline to produce a model that can adjust its diversity adaptively depending on the task at hand for general instruction following, and even use different decoding parameters within a single response for the best outcome.
Additionally, our proposed approach is general, and the \AD{} could be potentially used to convert hyperparameters other than temperature (e.g. top-p, top-k) effectively into model parameters. 
Furthermore, we show that \method{} is also a general tool to train  discrete latent variables that can be used for other architecture choices that contain discrete decisions.

\begin{figure*}[t]
    \centering
    \includegraphics[width=0.85\textwidth]{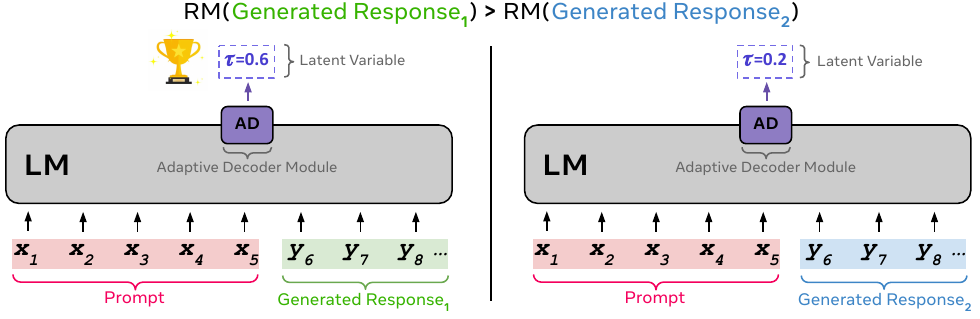}
    \caption{\textbf{\methodmedium{} (\method{}) Training Mechanism.} We demonstrate how preference pairs are constructed for training the \method{} loss (we show a Sequence-Level \AD{}, but the procedure remains the same for Token-Level). Here we have N=2 generated response samples for a single prompt, and the Reward Model (RM) scores Response$_1$ better than Response$_2$. Therefore, we use $\tau=0.6$ as the chosen temperature, and $\tau=0.2$ as the rejected temperature, and then apply the loss to prefer the chosen temperature over the rejected one for the given context (prompt).}
    \label{fig:ad_dpo_pairs}
\end{figure*}

\section{Related Work}

{\bf Fixed Decoding Strategies.}
Various methods have proposed different fixed decoding strategies that often depend on one or more hyperparameters. Beam search is a classical approach for tasks like machine translation \citep{freitag2017beam}, but is typically not used for LLM instruction following. A beam size of 1 corresponds to greedy decoding. \citet{holtzman2019curious} introduced nucleus sampling, \citet{fan2018hierarchical} introduced top-k sampling, and since then various further sampling approaches have been proposed.
\citet{shi2024thorough} showed that different decoding strategies work better for different tasks. \citet{zhang2020trading} evaluates different decoding strategies including fixed temperature, top-k, and top-p. They find that when diversity is the priority, all methods perform similarly, but when quality is the priority, top-p is the best. 
Using different temperatures to decode for different tasks has also cemented itself as common wisdom for prompting LLMs \citep{achiam2023gpt}. Commercial LLM API guides even recommend using a low temperature for analytical tasks and a temperature close to 1.0 for creative tasks \footnote{\url{https://docs.anthropic.com/en/api/complete}, \url{https://ai.google.dev/gemini-api/docs/text-generation}}. 

{\bf Adaptive Decoding.}
Prior work studied the adaptive change of decoding parameters under different criteria such as based on the target task, approximate reward of the desired output, or the target likelihood score.
\cite{zhu2024improving} developed a decoding strategy that can adapt based on the probability distribution of the previous token while \cite{zhu2023improving} uses a rule-based method to predict a temperature value for each token. \citet{basu2020mirostat} uses the desired perplexity value to predict the optimal top-k hyper-parameter for each token. \citet{finlayson2023closing} proposes basis-aware sampling that finds the optimal support over the next token distribution by addressing the softmax bottleneck issue.
Unlike our approach, none of these methods 
learn to predict an adaptive decoding strategy, but rather use various test time heuristics. \citet{li2024dynamic} propose a method to learn sample specific diversity values on dialogue tasks using an MSE loss, where the diversity values are then mapped to temperatures using a mapping function. \citet{zhang2024edt} dynamically select a temperature as a function of the entropy where the parameters of the function are treated as hyperparameters which they tune for each different task. 
Ad-hoc temperature prediction has been commonly used for calibration, as explored by \citet{kumar2019calibration} and \citet{xie2024calibrating}. 
To the best of our knowledge, we propose the first method to predict the temperature directly using preference optimization, allowing the model to learn task dependent temperatures at both the sequence and token levels.

{\bf Preference Optimization.}
Reinforcement Learning from Human Feedback (RLHF) has emerged as a major ingredient of LLM training \citep{ouyang2022training}.
DPO \citep{rafailov2024direct} and other preference optimization methods \citep{xu2023some,meng2024simpo} have significantly simplified the RLHF process. 
While many of these methods improve performance and generalization they can also negatively affect diversity and calibration \citep{achiam2023gpt,kirk2023understanding}.
In particular, RLHF methods optimize the final reward which does not take diversity into account, so it has become common practice to add a KL regularization term to maintain some of the model's original diversity \citep{ziegler2019fine,rafailov2024direct}.
To the best of our knowledge, our method is the first to use preference optimization for training latent variables instead of word tokens.

\section{Method}
The goal of our method is to make the language model itself choose an ideal temperature for generating tokens depending on the current context.
To achieve this, we add a small differentiable module to an existing LLM that predicts a temperature value to be used for decoding word tokens, which we call the \AD{}.
For training an \AD{} module, we develop a preference optimization method, \method{}, that is designed for learning such hyperparameters. In the following subsections we describe the \AD{} module and \method{} loss in more detail.

\subsection{\AD{} Module}
Here we introduce the \AD{} module, which is a small neural network that can be attached on top of any existing LLM.
It takes as an input latent representations of the last hidden layer and outputs a probability distribution over possible temperature choices.
Let $\mathcal{M}$ be a transformer core \citep{vaswani2017attention} within an LLM that maps a sequence of tokens $\{x_t\}$ to a latent representation, $h_t$, at the last layer. 
This latent representation is then usually converted to token probabilities using an un-embedding matrix $W$ followed by a softmax. 
Thus, a regular LLM generates the next token $x_{t+1}$ as follows:
\begin{equation}
\begin{gathered}
    h_{t} = \mathcal{M}(x_1, \dots x_{t}),  \\  x_{t+1} \sim \textsc{Softmax}(W h_{t}).
\end{gathered}
\end{equation}
A fixed temperature value, $\tau$, can be used to scale the softmax distribution in the following way:
\begin{equation}
\label{eq:sample_token}    
    x_{t+1} \sim \textsc{Softmax}(W h_{t}/\tau),
\end{equation}
where small temperature values (toward 0) make the distribution sharper, and high temperature values (toward 1) will result in the original distribution.

Adaptive Decoding works by predicting the optimal $\tau$ value for a specific input $\{x_t\}$. To add Adaptive Decoding to this LLM, we also feed the same latent representation $h_t$ to an \AD{} module that maps it to a categorical probability distribution over a set of pre-defined temperature values $\tau_1,\dots, \tau_K$:
\begin{equation}
\begin{gathered}
    P(\tau_k | h_t) = \textsc{AdaptiveDecoder}(h_t), \\
    \text{where} \enspace \sum_k P(\tau_k | h_t) = 1.
\end{gathered}
\end{equation}
We  can then straightforwardly make use of this distribution for generating a given output token, $x_{t+1}$, by selecting the temperature with the highest probability, and then use that temperature for decoding the next token:
\begin{equation}
    \label{eq:sample_temp_greedy}    
\begin{gathered}
    \tau = \text{argmax}_{\tau_k} P(\tau_k | h_t), \\ 
    x_{t+1} \sim \textsc{Softmax}(W h_{t}/\tau) .
\end{gathered}
\end{equation}
Alternatively, one can sample a \textit{temperature} from the distribution and then sample a \textit{token} with it:
\begin{equation}
    \label{eq:sample_temp}  
\begin{gathered}
    \tau \sim P(\tau_k | h_t), \\  
    x_{t+1} \sim \textsc{Softmax}(W h_{t}/\tau) .
\end{gathered}
\end{equation}
This latter approach can also be written as a single sampling operation:
\begin{equation}
\label{eq:temp_margin}    
    x_{t+1} \sim \sum_k P(\tau_k | h_t) \textsc{Softmax}(W h_{t}/\tau_k) .
\end{equation}
While the last two operations are identical, the second version will allow us to develop a new loss function for training as we will see in the next section.

Any neural network architecture can be used for the internals of the \AD{} module, but we use a multi-layer perceptron (MLP) with a softmax output for simplicity (details in ~\autoref{sec:setup}).
Note that it is also straightforward to generalize the \AD{} to other decoding hyperparameters such as top-k by simply modifying \autoref{eq:sample_token} to the corresponding operation.
In addition, $\mathcal{M}$ can be another neural model besides a transformer, such as a recurrent neural network.

\subsection{Token vs Sequence Level \AD{}.}

We propose two variants of the \AD, as demonstrated in ~\autoref{fig:AdaptiveDecoder}.
Let $\vec{x}=\{ x_1, \ldots, x_T \}$ be the sequence of given input prompt tokens, and $\vec{y}=\{ y_{T+1}, \ldots, y_{T'} \}$ be the generated response tokens.
In the token level variant, \ADtok{} (\smallADseq{}), a temperature is predicted for each new token to be decoded.
This is achieved by applying the \AD{} at every step of generation and using the selected temperature for sampling the following token:
\begin{equation}
\begin{gathered}
    \tau_t \sim  \textsc{AdaptiveDecoder}(h_t), \\ 
    y_{t+1} \sim \textsc{Softmax}(W h_{t}/\tau_t)
    \enspace \text{for} \enspace T \le t < {T'}.
\end{gathered}
\end{equation}
Such fine-grained temperature adjustment allows the model to learn an individual temperature value for each token. 

In the sequence level variant \ADseq{} (\smallADseq{}), a single temperature is predicted for the entire response. 
Unlike the token level, the \AD{} module is used only once per input prompt, applied at the last token $x_T$ of the prompt to predict a temperature value $\tau$ to be used for the entire response generation:
\begin{equation}
\begin{gathered}
    \tau \sim  \textsc{AdaptiveDecoder}(h_T), \\ 
    y_{t+1} \sim \textsc{Softmax}(W h_{t}/\tau)
    \enspace \text{for} \enspace T \le t < {T'}.
\end{gathered}
\end{equation}
Such a coarse-grained temperature adjustment may be sufficient for most applications where the task requires either conciseness or creativity, but not both, and is still potentially much more flexible than the classical approach of choosing a single fixed temperature for all input prompts.

\subsection{\methodlong{}}
\label{sec:struct_dpo}

To learn the \AD{} parameters, we employ a preference optimization training where we generate multiple responses from the model and label some of them as \emph{chosen} and others \emph{rejected}.
The overall goal of the training is to make the likelihood of generating chosen responses higher than the rejected ones, similar to the existing preference optimization methods such as DPO \citep{rafailov2024direct}.
However, those existing  methods are designed to train token probabilities, not latent variables within the model.
Thereby, we propose a generalization of DPO, which we call \methodlong{} (\method{}), that is a general approach  to train discrete latent variables, such as the choices of temperature\footnote{While temperature is a continuous value, we are focusing on discrete temperature options in this paper. This also makes it easy to generalize our method to other discrete variables, such as top-k.}. 

To use \method{} to learn optimal temperatures, we first generate multiple responses $\{\vec{y}^1, \ldots, \vec{y}^N\}$ for each prompt $\vec{x}$ by sampling temperatures from the \AD{} output, which then affect how tokens are sampled. 
Let $\vec{\tau}^n = \{\tau_{T+1}^n, \ldots, \tau_{T'}^n \}$ be the temperatures used when generating tokens in response $\vec{y}^n = \{y_{T+1}^n, \ldots, y_{T'}^n \}$.
The responses are then scored, either using an external reward model, or measuring the correctness of their answer, depending on the task.
The highest and lowest scoring responses become our chosen and rejected response pair $(\vec{y}^c, \vec{y}^r)$.
This process is depicted in \autoref{fig:ad_dpo_pairs}.
Regular DPO training would optimize the token probabilities of these response pairs, but our goal is to learn the corresponding chosen and rejected temperature values $(\vec{\tau}^c, \vec{\tau}^r)$ that are used when sampling the response tokens.
For this, there are multiple ways to adapt the DPO loss, which we outline below.

\textbf{Temperatures as tokens.}\,
The simplest formulation is to treat the temperature selection just like another token.
In this view, the model generates two tokens per step: a temperature token $\tau_t$ and a word token $y_t$.
The temperature tokens have a different vocabulary, consisting of possible temperature values, but that does not complicate training.
Similar to how the previous word token choice affects the next word token, the temperature token also affects the following word token probabilities.
Since the model is generating a single sequence of ``tokens'', $\vec{\hat{y}}^n = (\vec{y}^n, \vec{\tau}^n)$, we can apply the usual DPO loss to this joint token sequence:
\begin{align*}
    \mathcal{L}_\text{\method{}}&= -\log \sigma \left[\beta \log \frac{P(\vec{\hat{y}}^c )}{P_\text{ref}(\vec{\hat{y}}^c)}  - \beta \log \frac{P(\vec{\hat{y}}^r)}{P_\text{ref}(\vec{\hat{y}}^r)} \right] \notag \\ &= -\log \sigma \bigg[\beta \log \frac{P(\vec{y}^c , \vec{\tau}^c )}{P_\text{ref}(\vec{y}^c, \vec{\tau}^c)}  
    - \beta \log \frac{P(\vec{y}^r , \vec{\tau}^r )}{P_\text{ref}(\vec{y}^r, \vec{\tau}^r)} \bigg],
\end{align*}
where $P_\text{ref}$ are reference model probabilities. Since our reference model does not have an \AD{} module, we omit it for the temperature tokens\footnote{This is the same as assuming the reference model has always uniform probabilities over possible temperatures.}, and the loss therefore becomes:
\begin{multline}
\label{eq:loss1}
\mathcal{L}_\text{\method{}} = -\log \sigma \bigg[\beta \log \frac{P(\vec{y}^c)}{P_\text{ref}(\vec{y}^c)} - \beta \log \frac{P(\vec{y}^r)}{P_\text{ref}(\vec{y}^r)} \\ + \beta \log P(\vec{\tau}^c) - \beta \log P(\vec{\tau}^r) \bigg] .
\end{multline}
The advantage of this loss is that it takes into account both word token and temperature probabilities, making it possible to train both using a single loss.
Here $\beta$ is a hyperparameter of DPO that controls the KL term.

\textbf{Temperatures as tokens (separate).}\,
Like the previous formulation, we view the temperatures as tokens, but treat the word token generation as an external mechanism and focus only on the \AD{}.
In this view, the \AD{} module generates a token $\tau_t$, which is a temperature value in this case, that is then fed to an external mechanism that generates the word token $y_t$.
This framing makes things simpler because we have the \AD{} generating two sequences of temperature values $(\vec{\tau}^c, \vec{\tau}^r)$ where one is preferred over the other.
So we can directly apply the DPO loss with only the temperature tokens $\tau_t$:
\begin{equation}
\label{eq:loss2}    
\mathcal{L}_\text{\method{}} = -\log \sigma \left[ \beta \log P(\vec{\tau}^c) -  \beta \log P(\vec{\tau}^r) \right] .
\end{equation}
Again we omit the reference probabilities for the temperature tokens.
This loss is simple and does not take account of token probabilities, but one can also use a separate DPO loss for the word tokens.

\textbf{Temperatures as latents.}\,
In this version, we take advantage of the fact that the chosen and rejected labels are only conditioned on word tokens, and the temperature values that are used do not directly affect this ranking.
The real objective we want to optimize is the probability of sampling chosen and rejected word sequences.
Therefore, we treat the \AD{} as an internal mechanism of the model and the temperature values as latent variables.
This way, the model only outputs token probabilities like normal LLMs, but those probabilities are altered by the \AD{}, as follows:
\[
y_t \sim P'(y) = \sum_{\tau} P(y | \tau) P(\tau)  .
\]
Now we can apply the DPO loss to these token probabilities where the temperature is marginalized out
\begin{align}
\label{eq:loss3}  
\mathcal{L}_\text{\method{}} &= -\log \sigma \left[ \beta \log \frac{P'(\vec{y}^c)}{P'_\text{ref}(\vec{y}^c)} - \beta \log \frac{P'(\vec{y}^r)}
{P'_\text{ref}(\vec{y}^r)} \right] \notag \\
&= -\log \sigma \left[ \beta \sum_t \log \frac{P'({y}_t^c)}{P'_\text{ref}({y}_t^c)} 
- \beta \sum_t \log \frac{P'({y}_t^r)}
{P'_\text{ref}({y}_t^r)} \right] \notag \\
&= -\log \sigma \Bigg[ \beta \sum_t \log \frac{ \sum_\tau P({y}_t^c|\tau) P(\tau)}{\sum_\tau P_\text{ref}({y}_t^c|  \tau) P_\text{ref}(\tau)} \notag \\ & \hspace{40pt} - \beta \sum_t \log \frac{ \sum_\tau P({y}_t^r|\tau) P(\tau)}{\sum_\tau P_\text{ref}({y}_t^r|  \tau) P_\text{ref}(\tau)}\Bigg].
\end{align}
Note that the actual temperatures used in generation are irrelevant here, thus reducing the noise caused by sampling temperatures during training.
The reference temperature probabilities $P_\text{ref}(\tau)$ are uniform if that is the initialization.

\section{Experiments}

\begin{figure*}[t]
    \centering

    \includegraphics[width=0.48\textwidth]{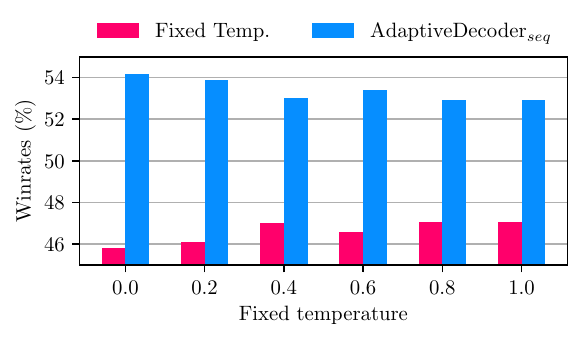}
    \hfill
    \includegraphics[width=0.48\textwidth]{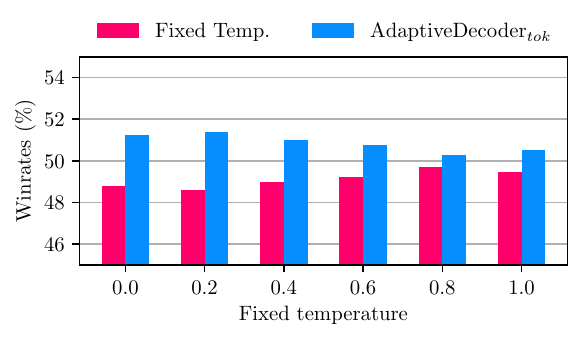}
    \caption{\textbf{UltraMathStories Results.} UltraMathStories is a superset of UltraFeedback, GSM8K, and Stories. The {Adaptive Decoding} models are trained on all 3 subtasks simultaneously. Winrates are shown as the average winrate across the test sets of the 3 subtasks in UltraMathStories. \textbf{(left)} \ADseq{} vs Fixed Temperature Winrates. \textbf{(right)} \ADtok{} vs Fixed Temperature Winrates. In both cases, Adaptive Decoding outperforms all fixed temperatures.}
    \label{fig:ultramathstories}
\end{figure*}

\begin{table*}[]
\centering
\begin{tabular}{lcc}
\toprule
\bf Method & \bf 3-gram-repeats $\downarrow$ & \bf \% of non-greedy \\
\midrule
Greedy Decoding  & 0.36\%
 & 0\% \\
\ADtok{} & \textbf{0.22\%} & 94\% \\
\bottomrule
\end{tabular}
\label{tab:repeats}
\caption{\textbf{Reducing Repeats using the \AD{}.} We feed text from Wikitext-2 to the model and ask it to complete it. When completing a text, \ADtok{} learns to avoid greedy decoding in order to reduce repeats. In 94\% of samples, \ADtok{} learns to pick a non-greedy temperature. }
\label{tab:repeats}
\end{table*}

\subsection{Setup}
\label{sec:setup}
For all experiments, we train an \AD{} on top of a Llama 3.0-8B-Instruct model \citep{dubey2024llama}. 
The \AD{} module is a 3-layer MLP with hidden dimension 2048, and SiLU \citep{hendrycks2016gaussian} activations. 
We freeze the weights of the Llama model to better understand the effect of sampling temperature in isolation from finetuning the whole model.
For \method{} training, by default we use the loss in \autoref{eq:loss2} for its simplicity, unless otherwise specified.
During training, responses are generated using  \autoref{eq:sample_temp} where temperatures are sampled, but we use greedy temperature selection at inference time using \autoref{eq:sample_temp_greedy}, unless otherwise specified.

\subsection{Reducing N-gram Repetitions}
We start with a simple first experimentwhere we know temperature choice matters. 
It is understood that LLMs are prone to erroneous repetitions, particularly when greedy decoding ($\tau$=0) is used \cite{holtzman2019curious}.
We therefore sought to validate whether the \AD{} can learn to pick higher temperatures for specific tokens to avoid repeats. 
We use an \ADtok{} and provide it with 5 temperature options: $\tau \in \{0.0, 0.1, 0.2, 0.4, 0.6\}$.
We feed text from Wikitext-2 \citep{merity2016pointer} to the model and ask it to complete it.
We use 3-gram-repeats to rank the responses and create preference training pairs (see
\autoref{sec:ultramathstories_details} for details).
We find that the \ADtok{} effectively learns to reduce repeats by 42\% compared to greedy decoding on the Wikitext-2 test set (\autoref{tab:repeats}). 
We also note that in around 94\% of cases \ADtok{} learns to pick a non-greedy temperature. 
This serves as a proof of concept that \method{} can successfully optimize the temperature values in the right direction at the token level.

\begin{figure*}[t]
  \centering
  \begin{subfigure}[b]{0.31\textwidth}
    \includegraphics[width=\textwidth]{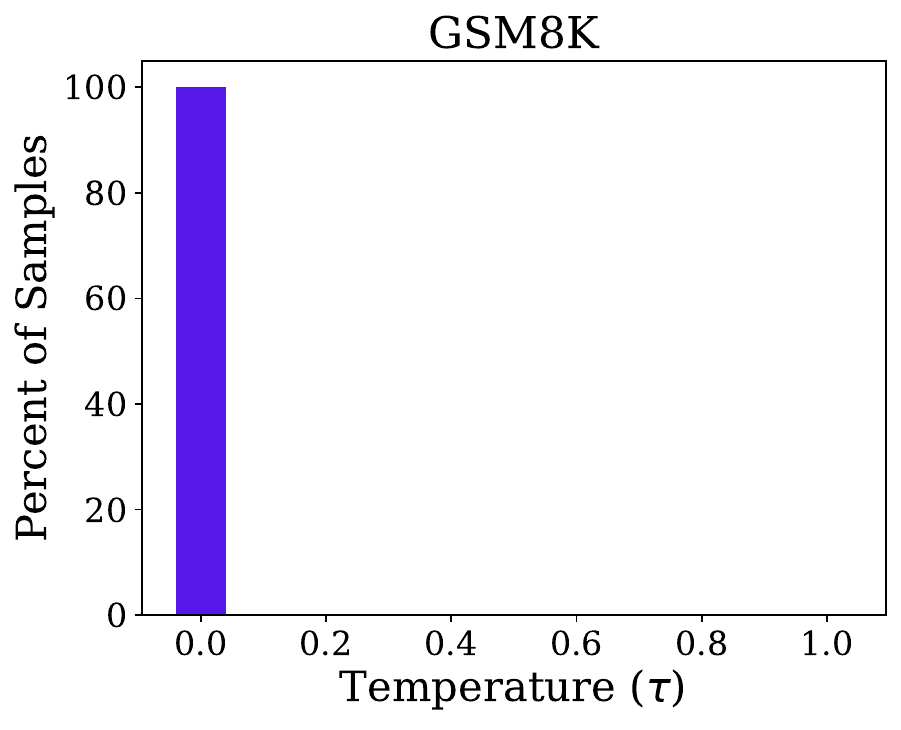}
  \end{subfigure}
  \hfill
  \begin{subfigure}[b]{0.305\textwidth}
    \includegraphics[width=\textwidth]{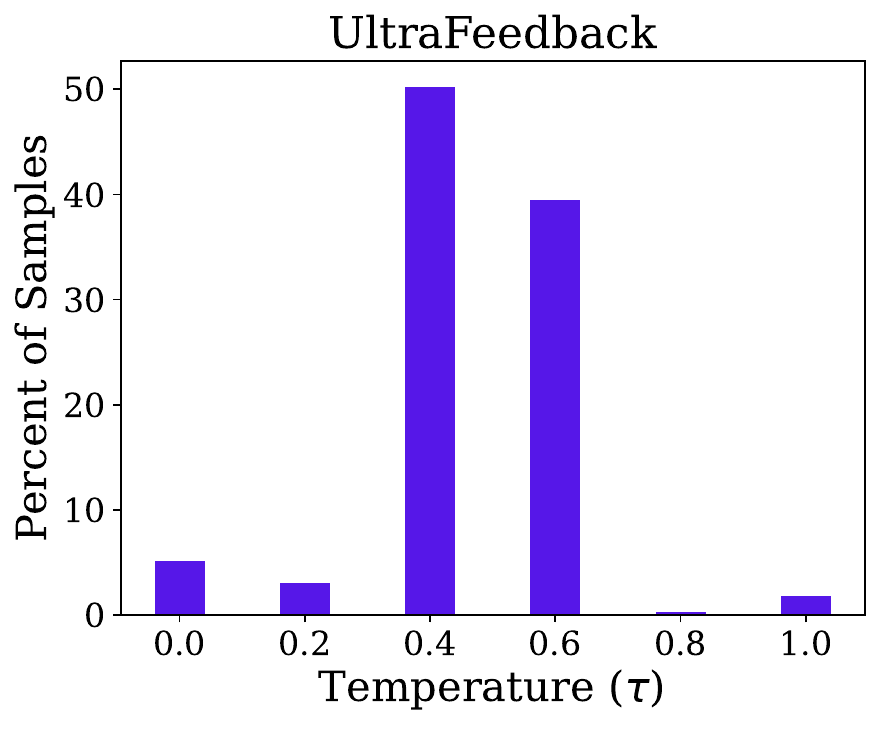}
  \end{subfigure}
  \hfill
  \begin{subfigure}[b]{0.31\textwidth}
    \includegraphics[width=\textwidth]{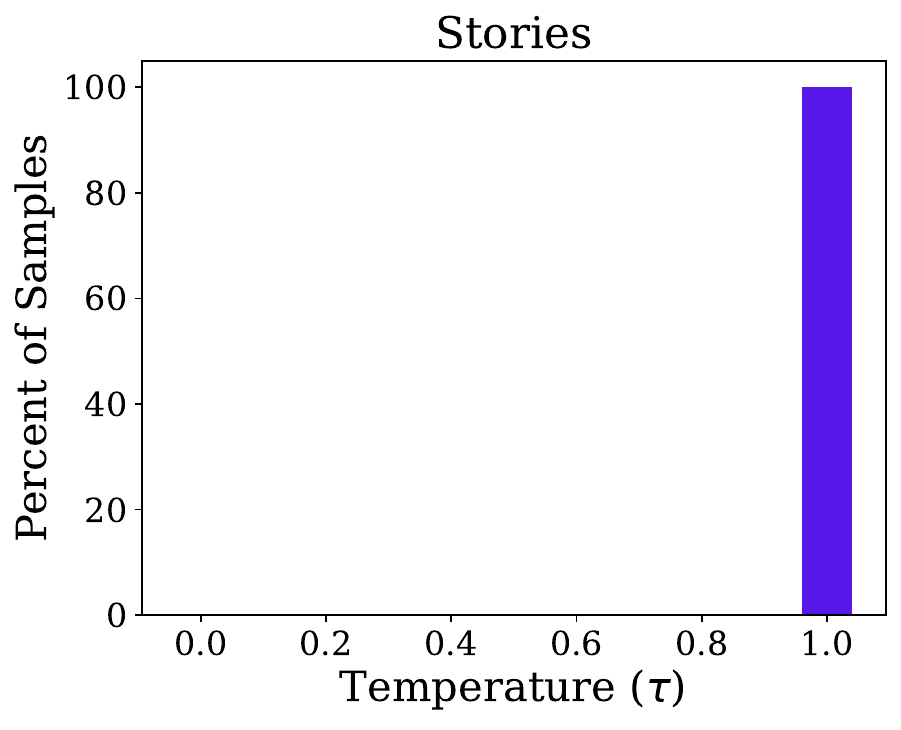}
  \end{subfigure}
  \caption{\textbf{\ADseq{} predicted temperature distributions.} We show the distribution of predicted temperatures on the test set of each subtask in UltraMathStories. As expected, the model predicts low temperatures for GSM8K, high temperatures for Stories, and temperatures mostly in between for UltraFeedback.}
  \label{fig:ad_seq_temp_distributions}
\end{figure*}

\begin{table*}[t]
\centering
\begin{tabular}{p{14.5cm}c}
\toprule
{\bf Prompt} & {\bf Predicted $\tau$} \\
\midrule \small
Detailed Instructions: In this task, you are given a country name and you need to return the capital city of the given country\textbackslash n
Problem:Guinea-Bissau\textbackslash n
Solution: &
 0.0 \\
\midrule
Write a compelling short story about a bitter and intense rivalry between two individuals, where one must have an advantage in terms of their socioeconomic status or physical ability. The story must also incorporate a surprising twist that leads to an unforeseen outcome. & 1.0\\
\bottomrule

\end{tabular}
\caption{Examples of \textbf{\ADseq{} Predicted Temperatures ($\tau$) on UltraFeedback.}
Here we show examples of UltraFeedback test prompts where the \ADseq{} model predicted $\tau \in \{0.0, 1.0\}$. That is, our model predicts the top prompt requires a factual deterministic response ($\tau=0.0$), while the bottom prompt requires a  creative, stochastic response ($\tau=1.0$).
More examples are shown in Appendix \autoref{tab:ultrafeedback_examples_appendix}.
}
\label{tab:ultrafeedback_examples_main}
\end{table*}

\subsection{UltraMathStories}
\label{sec:ad_seq_ultramathstories}
Next, we consider a much more realistic general instruction following setting 
to test if the \AD{} can learn to select different temperatures depending on the given prompt query. We thus deliberately consider a dataset that is a mixture of the following subtasks that require both formulaic, as well as creative responses:
\begin{itemize}[leftmargin=0.3cm,itemindent=0.2cm]
\item \textbf{Math (GSM8K).}
When solving math reasoning problems, LLMs require greedy, or low-temperature sampling to produce accurate and reliable results \citep{kojima2022large}.
The model should not deviate from high-likelihood tokens in this setting since factuality is crucial for finding the correct answer. 
GSM8K \citep{cobbe2021training} is a common math reasoning dataset used to evaluate such capabilities. 
Since we have the ground truth answers, we use them to score responses to select training pairs, and for final test evaluation.

\item \textbf{Creative Writing (Stories).}
In contrast, when solving open-ended creative writing problems, LLMs benefit from high temperature sampling to write more interesting and original responses. 
We introduce a creative story writing task, which we call ``Stories'', to evaluate the creativity and coherence of a model on open ended prompts.
We prompt the model to write a short story of a given title, where we use a language model to create the initial task titles. We use the Armo reward model (ArmoRM) \citep{wang2024interpretable} for scoring responses and selecting training preference pairs, as well as for scoring responses during final evaluation. ArmoRM gives a scalar score value for a response given its prompt. We take the top and bottom scored responses for selecting the chosen/rejected pairs See Appendix \autoref{app:stories} for more details on creating the dataset, and constructing the preference pairs.

\item \textbf{General Instructions (UltraFeedback).} Finally, many real-world prompts lie somewhere in between structured reasoning and open-ended creativity, or contain a mixture of both. 
We therefore consider the UltraFeedback \citep{cui2023ultrafeedback} dataset, 
which covers a wide variety of real user prompts, ranging from rigid reasoning tasks to open-ended writing. 
We use the same ArmoRM for constructing training preference pairs and evaluating.
\end{itemize}
We combine 2,000 training samples from UltraFeedback, 1,000 training samples from GSM8K, and 1,000 training samples from the Stories dataset, and call it the ``UltraMathStories'' dataset. We train a single model on the preference pairs from this dataset to test if {\em Adaptive Decoding} can adapt to each subtask. 
We evaluate on each subtask's test set individually and take the average winrate across the 3 test sets. 
Further details of each subtask included in this dataset, including how the \method{} pairs are created, are described in \autoref{sec:ultramathstories_details}. 
We experiment with both a sequence level and token level \AD{}, and provide each with 6 temperature options: $\tau \in \{0.0, 0.2, 0.4, 0.6, 0.8, 1.0\}$.
\\

\textbf{\AD{} can learn to use the ideal temperature adapted for each subtask.}
In \autoref{fig:ultramathstories}, we directly compare our method against fixed temperature decoding.
The winrate in each subtask is computed (shown in \autoref{sec:ad_seq_winrates}) and their average is plotted.
We observe the \AD{} outperforming all of the fixed temperatures, which indicates that the decoder has learned to choose an ideal decoding temperature suited to each subtask.
In fact, \autoref{fig:ad_seq_temp_distributions} demonstrates this clearly with the predicted temperature distributions for each subtask. 
As expected, the \AD{} predicts low temperature for math prompts (GSM8K), high temperature for creative writing prompts (Stories), and a mix of temperatures which are mostly in between for general instruction prompts (UltraFeedback).
The latter has the biggest temperature variance, which makes sense given that it has more diverse prompts.

\textbf{Sequence-level vs.\ Token-level \AD{}.}
In this task, \ADseq{} showed a stronger performance compared to \ADtok{} as shown in \autoref{fig:ultramathstories}, even though both outperform fixed temperatures.
There are several reasons why this can be the case.
First, the subtasks in UltraMathStories themselves might not require fine-grained temperature adjustment.
Secondly, learning a single temperature value per sample is much easier, thus likely to require fewer training samples (we only train on 4000 samples in total).
However, we will explore the advantage of \ADtok{} in subsequent sections.

\subsection{Constrained Creative Writing (ConstrainedStories)}

\begin{figure*}[t]
  \centering
  \begin{subfigure}[b]{0.49\textwidth}
    \includegraphics[width=\textwidth]{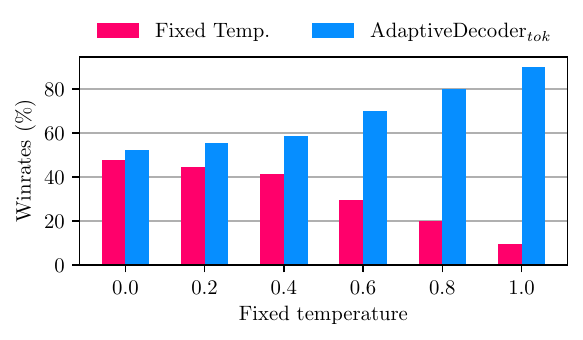}
  \end{subfigure}
  \hfill
  \begin{subfigure}[b]{0.46\textwidth}
    \includegraphics[width=\textwidth]{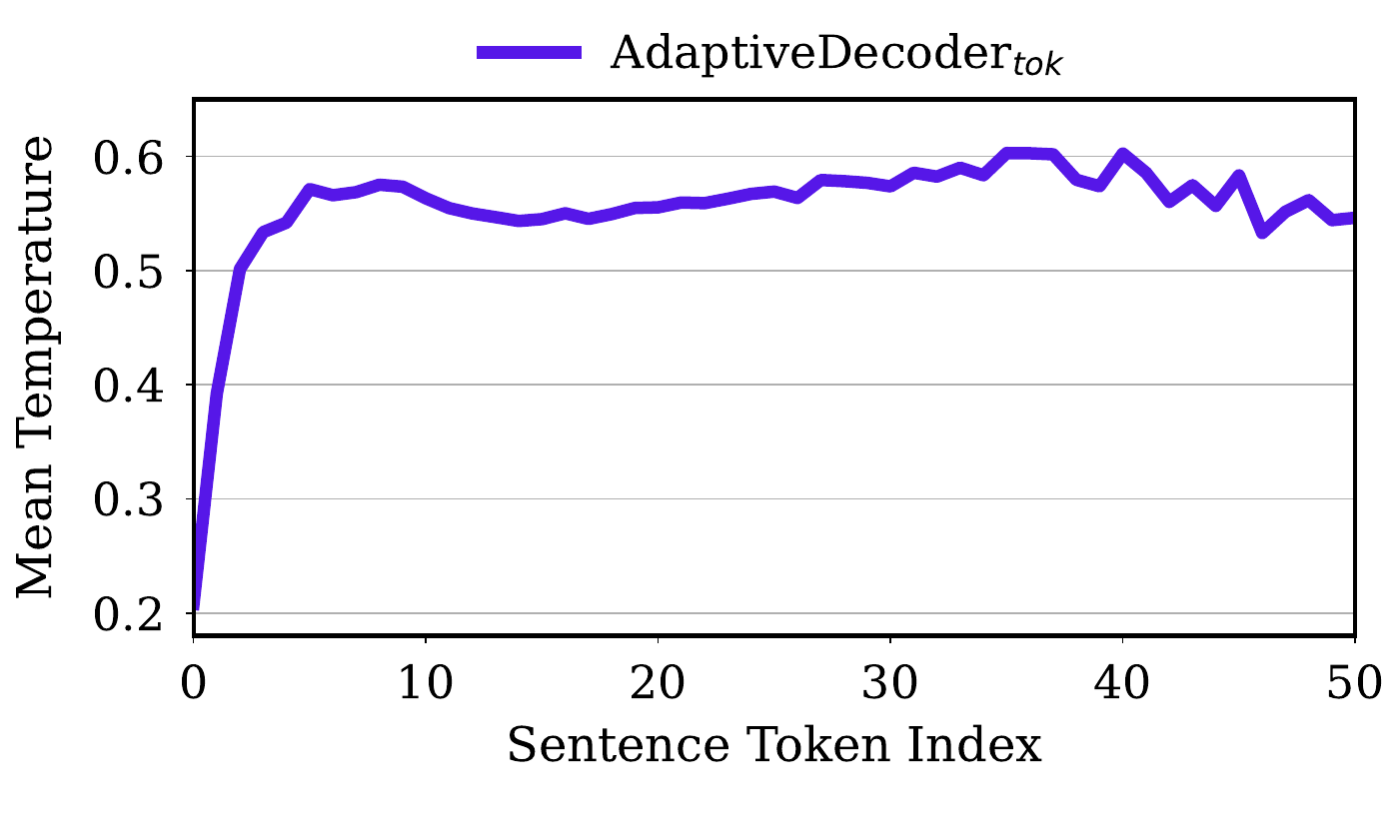}
  \end{subfigure}
  \caption{\textbf{Constrained Creative Writing (ConstrainedStories) Results.} Here we show a quantitative analysis of the \AD{} on the constrained creative writing task, ConstrainedStories. \textbf{(left)} \ADtok{} winrates vs fixed temperatures. 
  The high fixed temperatures perform worse because they fail to follow the constraint. Fixed greedy decoding works well at following the constraint, but \ADtok{} outperforms it by using higher temperatures when possible. \textbf{(right)} Mean temperature predicted by the \ADtok{} for the first 50 tokens of each sentence. This plot confirms our hypothesis that the first token of each sentence should be low temperature in order to follow the constraint, and all other tokens should be high temperature in order to write a good story. The average temperature for the first token is $\tau=0.21$, and the average temperature for all other tokens is $\tau=0.55$, showing a more greedy decoding for the constraint, and less greedy everywhere else.}
  \label{fig:constrained_creative_writing}
\end{figure*}
\begin{table*}[t]
    \centering
\scalebox{0.9}{
\begin{tabular}{lcc}
\toprule
\bf Decoding Method & \makecell{\bf Accuracy $\uparrow$ \\ (Majority of N=8 responses)} & \makecell{\bf Accuracy $\uparrow$ \\ (N=1 response)} \\
\midrule
Best Fixed Temperature & 87.46 & 81.59\\
\ADtok{}  ($\tau \in \{0.0, 0.4, 0.8, 1.0\}$) & 87.70  & 80.47 \\
\ADtok{} ($\tau \in \{0.0, 0.4, 0.8, 1.0, 1.2\}$)  & {\bf 87.95} & 80.51\\
\bottomrule
\end{tabular}
}
    \caption{\textbf{\ADtok{} for majority voting (8 samples) on the GSM8K dataset.} \ADtok{} learns to assign appropriate temperatures at different parts of the generation which allows for more accurate sampled reasoning chains which results in a higher accuracy than using a single tuned temperature for the dataset. We also include the accuracy for N=1 response, which underperforms majority voting.}
    \label{tab:maj_vote}
\end{table*}

When given rigid instructions such as solving a math problem, the model needs to be greedy, but when given open-ended instructions such as writing a creative story, the model needs to be non-greedy. However, certain instructions can contain both rigid and open-ended instructions. We consider the problem of constrained creative writing, which requires the model to be both greedy and non-greedy at different tokens of a single response.

We construct a dataset based on the Stories dataset from the previous subsection, and call it ``ConstrainedStories''. Similar to the Stories task, we prompt the model to write a creative story of a given title, but with an extra instruction saying that each sentence must start with a specific substring, ``Ab'' in this case.
Intuitively, one would expect the ideal model should be greedy when generating the start of each sentence to satisfy the constraint, and non-greedy everywhere else for better creativity.
The \method{} preference pairs are created using both ArmoRM scores and constraint satisfaction rates.
During evaluation, a higher constraint satisfaction wins, but ties are broken by the ArmoRM score.
More details can be found in Appendix \autoref{app:constrainedstories}.

\textbf{\ADtok{} can learn to dynamically adjust the temperature at the token-level. } \autoref{fig:constrained_creative_writing} (left) shows the winrates of \ADtok{} compared to fixed temperature decoding. The \ADtok{} always outperforms fixed temperature decoding.
When a high fixed temperature is used on all tokens, it fails to follow the constraint, resulting in a low winrate.
The greedy decoding performs well as it satisfies the constraint more often, but the story quality is lowered by the lack of diversity. \autoref{tab:constrained_full_winrates} shows the individual winrates for constraint satisfaction and Armo score.
As shown in \autoref{fig:constrained_creative_writing} (right), the \ADtok{} manages to have the best of both worlds.
The average temperature for the first token of each sentence is $\tau=0.21$, and the average temperature for all other tokens is $\tau=0.55$. This shows that the model is mostly greedy on the constraint tokens (in order to generate an ``Ab'' word at the start of each sentence), and mostly non-greedy on all other tokens (in order to generate a creative and coherent story). Appendix \autoref{fig:constrained_story_ex} shows an example of the \ADtok{} predicted temperatures for a test sample prompt in this task.

\begin{table*}[t]
\centering
\setlength{\tabcolsep}{8pt}
\begin{tabular}{cccccc}
\toprule
\multicolumn{3}{c}{\bf Fixed Temperature} & \multicolumn{3}{c}{\bf  \ADseq{}}\\
\cmidrule(r){1-3} \cmidrule(l){4-6}
\bf $\tau=0$ & $\tau=0.6$ & $\tau=1.0$ & 
\makecell{\bf $\mathcal{L}_\text{\method{}}$ \\ (\autoref{eq:loss2})} & \makecell{\bf $\mathcal{L}_\text{\method{}}$ \\ (\autoref{eq:loss3})}  & \makecell{\bf$\mathcal{L}_\text{NLL}$}\\
\midrule
\bf 81.59 & 
79.15 & 78.32 & 
\bf 81.59 & \bf 81.59 & 78.92 \\
\bottomrule
\end{tabular}
\caption{\textbf{GSM8K Accuracy comparing different loss functions for training a sequence-leval \AD{} (\smallADseq{}).} We compare two different $\mathcal{L}_\text{\method{}}$ loss functions, as outlined in \autoref{sec:struct_dpo}, as well as negative log likelihood loss, $\mathcal{L}_\text{NLL}$, trained on the chosen responses from the preference pairs.}
\label{tab:loss}
\end{table*}
\begin{table*}[t]
\centering
\begin{tabular}{llcc}
\toprule
& & \multicolumn{2}{c}{\bf Temperature Selection} \\
\cmidrule{3-4}
& & \makecell{ Greedy (\autoref{eq:sample_temp_greedy})} & \makecell{Sample (\autoref{eq:sample_temp})}  \\
\midrule
\multirow{6}{1.5cm}{\bf Fixed Temp.} & 
$\tau=0.0$ & {53.10} & {52.80} \\
& $\tau=0.2$ & {53.35} & {53.15} \\
& $\tau=0.4$ & {50.80} & {51.75} \\
& $\tau=0.6$ & {52.15} & {52.50} \\
& $\tau=0.8$ & {52.78} & {53.65} \\
& $\tau=1.0$ & {54.89} & {53.95} \\
\bottomrule
\end{tabular}
\caption{\textbf{\AD{} Temperature Selection Methods on UltraFeedback}. The \AD{} outputs a distribution over temperature values $\tau$, so we can either sample $\tau$ from that distribution or greedily select the highest probability $\tau$.
Here we show winrates against the fixed temperature decoding in the left column, using the \ADseq{} model trained on UltraMathStories (\autoref{sec:ad_seq_ultramathstories}).
All the winrates are above 50\%, which means the \AD{} always outperforms the fixed temperature. 
Also, we do not observe a significant difference between the two temperature selection methods.
}
\label{tab:ad_temp}
\end{table*}

\subsection{Majority Voting}

\citet{wang2022self} propose self-consistency, a method to improve the reliability of answers generated by language models by generating $N$ multiple independent reasoning chains and selecting the answer that appears most frequently.
When performing majority voting, several reasoning chains can be sampled (using a higher temperature) from the model, after which the most frequent answer is chosen. 
While greedy decoding is empirically found to be optimal for single response accuracy, obviously self-consistency cannot benefit from it as all of the generated responses will be identical. Therefore, it is important to find the ideal temperatures to use for generating the $N$ different responses per prompt.

Specifically, we explore whether the \AD{} can learn to ascertain which parts of the reasoning chain should be sampled more stochastically and which should be decoded greedily. 
We first train a \ADtok{} model on GSM8K to optimize the single response accuracy.
We do this by sampling $N=8$ responses and creating preference pairs using the ground-truth answers provided, and apply \method{} (\autoref{eq:loss2}). 
Then we evaluate this model in a majority voting setting and compare to the best fixed temperature decoding (tuned on the train set).
We experiment with two categories of possible temperatures: $\{0.0, 0.4, 0.8, 1.0\}$ and $\{0.0, 0.4, 0.8, 1.0, 1.2\}$.

Generally, we find that increasing the fixed temperature above 1.0 can cause the LLM's generation to start to degrade and this can also hurt the performance of majority voting. 
However, the \ADtok{} learns to assign temperatures appropriately and we observe that the higher temperature options help the model's performance, as shown in \autoref{tab:maj_vote}.
This demonstrates that the \ADtok{} trained by \method{} can result in a model that can perform well on both single responses (see \autoref{tab:loss} for single response accuracy) and majority voting setups at the same time.

\subsection{Ablations}

\subsubsection{\method{} Loss Type}

As described in \autoref{sec:struct_dpo}, there are several variations of the \method{} loss that we can use. Here we compare two different \method{} variants on the GSM8K math reasoning task: temperatures as tokens (separate) (\autoref{eq:loss2}) and temperatures as latents (\autoref{eq:loss3}). \autoref{tab:loss} shows the winrates of the \ADseq{} model trained with the two different losses on the GSM8K math reasoning task.
We see that both losses work well and match the greedy decoding (optimal) baseline. We also compare to a negative log-likelihood loss ($\mathcal{L}_\text{NLL}$), which is trained on only the chosen responses. This performs worse than both \method{} losses since it tends to predict the most frequently chosen temperature, which is not necessarily the best temperature, as demonstrated in the training sample distribution plots in Appendix ~\autoref{fig:training_distribitions}.

\subsubsection{\AD{} Temperature Selection}
The objective of the \AD{} is to predict the best temperature to use for a particular sample or token. This temperature is then used to scale the token probabilities for sampling a token. However, the \method{} training learns a distribution of temperatures, not just a single value. Therefore, at inference time we can either greedily select the top temperature as in \autoref{eq:sample_temp_greedy}, or sample from the temperature distribution following \autoref{eq:sample_temp}, as we do for sampling from the token distribution. Here we compare these two different ways of selecting temperatures. \autoref{tab:ad_temp} shows the winrates on UltraFeedback of the \ADseq{} model trained on UltraMathStories (\autoref{sec:ad_seq_ultramathstories}).
Both methods outperform all fixed temperature decoding temperatures, and we see a marginal difference between the two sampling methods.

\section{Conclusion}

As large language models continue to advance, users are still left with important hyperparameter choices when working with them on end use applications. Notably, decoding temperature is a crucial parameter at inference time for determining how much the model should explore (generate novel, creative text) vs exploit (generate conventional, factual text).

In this paper, we introduce the \AD{}, a neural module that sits on top of a pretrained LLM to predict what temperature should be used to sample the next token. The \AD{} is trained with our proposed \methodlong{} (\method{}) method, which can  optimize discrete latent variables such as temperature.
We find that across a variety of tasks, our method outperforms all fixed temperature values, eliminating the need for users to select a fixed temperature in advance, or tuning the right temperature for each task at evaluation time.
Our experiments demonstrate the effectiveness of training an \AD{} module on top of an existing (frozen) language model, making our approach usable and simple to employ with existing language models, as well as being used when developing new ones. 

Finally, Adaptive Decoding is a general approach, and the \AD{} could be potentially used to convert other hyperparameters than temperature effectively into training parameters.  \method{} is also a general tool to train  discrete latent variables that could similarly be used for other hyperparameters such as top-p or top-k. Our approach also opens the possibility of defining and exploring larger numbers of decoding hyperparameters, as they now can be trained rather than be manually set.

\bibliography{icml2022}

\begin{thebibliography}{33}
\providecommand{\natexlab}[1]{#1}
\providecommand{\url}[1]{\texttt{#1}}
\expandafter\ifx\csname urlstyle\endcsname\relax
  \providecommand{\doi}[1]{doi: #1}\else
  \providecommand{\doi}{doi: \begingroup \urlstyle{rm}\Url}\fi

\bibitem[Achiam et~al.(2023)Achiam, Adler, Agarwal, Ahmad, Akkaya, Aleman, Almeida, Altenschmidt, Altman, Anadkat, et~al.]{achiam2023gpt}
Josh Achiam, Steven Adler, Sandhini Agarwal, Lama Ahmad, Ilge Akkaya, Florencia~Leoni Aleman, Diogo Almeida, Janko Altenschmidt, Sam Altman, Shyamal Anadkat, et~al.
\newblock Gpt-4 technical report.
\newblock \emph{arXiv preprint arXiv:2303.08774}, 2023.

\bibitem[Ackley et~al.(1985)Ackley, Hinton, and Sejnowski]{ackley1985learning}
David~H Ackley, Geoffrey~E Hinton, and Terrence~J Sejnowski.
\newblock A learning algorithm for boltzmann machines.
\newblock \emph{Cognitive science}, 9\penalty0 (1):\penalty0 147--169, 1985.

\bibitem[Basu et~al.(2020)Basu, Ramachandran, Keskar, and Varshney]{basu2020mirostat}
Sourya Basu, Govardana~Sachitanandam Ramachandran, Nitish~Shirish Keskar, and Lav~R. Varshney.
\newblock Mirostat: A neural text decoding algorithm that directly controls perplexity, 2020.

\bibitem[Cobbe et~al.(2021)Cobbe, Kosaraju, Bavarian, Chen, Jun, Kaiser, Plappert, Tworek, Hilton, Nakano, et~al.]{cobbe2021training}
Karl Cobbe, Vineet Kosaraju, Mohammad Bavarian, Mark Chen, Heewoo Jun, Lukasz Kaiser, Matthias Plappert, Jerry Tworek, Jacob Hilton, Reiichiro Nakano, et~al.
\newblock Training verifiers to solve math word problems.
\newblock \emph{arXiv preprint arXiv:2110.14168}, 2021.

\bibitem[Cui et~al.(2023)Cui, Yuan, Ding, Yao, Zhu, Ni, Xie, Liu, and Sun]{cui2023ultrafeedback}
Ganqu Cui, Lifan Yuan, Ning Ding, Guanming Yao, Wei Zhu, Yuan Ni, Guotong Xie, Zhiyuan Liu, and Maosong Sun.
\newblock Ultrafeedback: Boosting language models with high-quality feedback.
\newblock \emph{arXiv preprint arXiv:2310.01377}, 2023.

\bibitem[Dubey et~al.(2024)Dubey, Jauhri, Pandey, Kadian, Al-Dahle, Letman, Mathur, Schelten, Yang, Fan, et~al.]{dubey2024llama}
Abhimanyu Dubey, Abhinav Jauhri, Abhinav Pandey, Abhishek Kadian, Ahmad Al-Dahle, Aiesha Letman, Akhil Mathur, Alan Schelten, Amy Yang, Angela Fan, et~al.
\newblock The llama 3 herd of models.
\newblock \emph{arXiv preprint arXiv:2407.21783}, 2024.

\bibitem[Fan et~al.(2018)Fan, Lewis, and Dauphin]{fan2018hierarchical}
Angela Fan, Mike Lewis, and Yann Dauphin.
\newblock Hierarchical neural story generation.
\newblock \emph{arXiv preprint arXiv:1805.04833}, 2018.

\bibitem[Finlayson et~al.(2023)Finlayson, Hewitt, Koller, Swayamdipta, and Sabharwal]{finlayson2023closing}
Matthew Finlayson, John Hewitt, Alexander Koller, Swabha Swayamdipta, and Ashish Sabharwal.
\newblock Closing the curious case of neural text degeneration, 2023.

\bibitem[Freitag \& Al-Onaizan(2017)Freitag and Al-Onaizan]{freitag2017beam}
Markus Freitag and Yaser Al-Onaizan.
\newblock Beam search strategies for neural machine translation.
\newblock \emph{arXiv preprint arXiv:1702.01806}, 2017.

\bibitem[Hendrycks \& Gimpel(2016)Hendrycks and Gimpel]{hendrycks2016gaussian}
Dan Hendrycks and Kevin Gimpel.
\newblock Gaussian error linear units (gelus).
\newblock \emph{arXiv preprint arXiv:1606.08415}, 2016.

\bibitem[Holtzman et~al.(2019)Holtzman, Buys, Du, Forbes, and Choi]{holtzman2019curious}
Ari Holtzman, Jan Buys, Li~Du, Maxwell Forbes, and Yejin Choi.
\newblock The curious case of neural text degeneration.
\newblock \emph{arXiv preprint arXiv:1904.09751}, 2019.

\bibitem[Kirk et~al.(2023)Kirk, Mediratta, Nalmpantis, Luketina, Hambro, Grefenstette, and Raileanu]{kirk2023understanding}
Robert Kirk, Ishita Mediratta, Christoforos Nalmpantis, Jelena Luketina, Eric Hambro, Edward Grefenstette, and Roberta Raileanu.
\newblock Understanding the effects of rlhf on llm generalisation and diversity.
\newblock \emph{arXiv preprint arXiv:2310.06452}, 2023.

\bibitem[Kojima et~al.(2022)Kojima, Gu, Reid, Matsuo, and Iwasawa]{kojima2022large}
Takeshi Kojima, Shixiang~Shane Gu, Machel Reid, Yutaka Matsuo, and Yusuke Iwasawa.
\newblock Large language models are zero-shot reasoners.
\newblock \emph{Advances in neural information processing systems}, 35:\penalty0 22199--22213, 2022.

\bibitem[Kumar \& Sarawagi(2019)Kumar and Sarawagi]{kumar2019calibration}
Aviral Kumar and Sunita Sarawagi.
\newblock Calibration of encoder decoder models for neural machine translation.
\newblock \emph{arXiv preprint arXiv:1903.00802}, 2019.

\bibitem[Li et~al.(2024)Li, Mi, Li, Wang, Sun, Feng, and Li]{li2024dynamic}
Yiwei Li, Fei Mi, Yitong Li, Yasheng Wang, Bin Sun, Shaoxiong Feng, and Kan Li.
\newblock Dynamic stochastic decoding strategy for open-domain dialogue generation.
\newblock \emph{arXiv preprint arXiv:2406.07850}, 2024.

\bibitem[Meng et~al.(2024)Meng, Xia, and Chen]{meng2024simpo}
Yu~Meng, Mengzhou Xia, and Danqi Chen.
\newblock Simpo: Simple preference optimization with a reference-free reward.
\newblock \emph{arXiv preprint arXiv:2405.14734}, 2024.

\bibitem[Merity et~al.(2016)Merity, Xiong, Bradbury, and Socher]{merity2016pointer}
Stephen Merity, Caiming Xiong, James Bradbury, and Richard Socher.
\newblock Pointer sentinel mixture models.
\newblock \emph{arXiv preprint arXiv:1609.07843}, 2016.

\bibitem[OpenAI(2023)]{openai2023gpt4}
OpenAI.
\newblock Gpt-4 technical report, 2023.

\bibitem[Ouyang et~al.(2022)Ouyang, Wu, Jiang, Almeida, Wainwright, Mishkin, Zhang, Agarwal, Slama, Ray, et~al.]{ouyang2022training}
Long Ouyang, Jeffrey Wu, Xu~Jiang, Diogo Almeida, Carroll Wainwright, Pamela Mishkin, Chong Zhang, Sandhini Agarwal, Katarina Slama, Alex Ray, et~al.
\newblock Training language models to follow instructions with human feedback.
\newblock \emph{Advances in neural information processing systems}, 35:\penalty0 27730--27744, 2022.

\bibitem[Rafailov et~al.(2024)Rafailov, Sharma, Mitchell, Manning, Ermon, and Finn]{rafailov2024direct}
Rafael Rafailov, Archit Sharma, Eric Mitchell, Christopher~D Manning, Stefano Ermon, and Chelsea Finn.
\newblock Direct preference optimization: Your language model is secretly a reward model.
\newblock \emph{Advances in Neural Information Processing Systems}, 36, 2024.

\bibitem[Shi et~al.(2024)Shi, Yang, Cai, Zhang, Wang, Yang, and Lam]{shi2024thorough}
Chufan Shi, Haoran Yang, Deng Cai, Zhisong Zhang, Yifan Wang, Yujiu Yang, and Wai Lam.
\newblock A thorough examination of decoding methods in the era of llms.
\newblock \emph{arXiv preprint arXiv:2402.06925}, 2024.

\bibitem[Vaswani(2017)]{vaswani2017attention}
A~Vaswani.
\newblock Attention is all you need.
\newblock \emph{Advances in Neural Information Processing Systems}, 2017.

\bibitem[Wang et~al.(2024)Wang, Xiong, Xie, Zhao, and Zhang]{wang2024interpretable}
Haoxiang Wang, Wei Xiong, Tengyang Xie, Han Zhao, and Tong Zhang.
\newblock Interpretable preferences via multi-objective reward modeling and mixture-of-experts.
\newblock \emph{arXiv preprint arXiv:2406.12845}, 2024.

\bibitem[Wang et~al.(2022)Wang, Wei, Schuurmans, Le, Chi, Narang, Chowdhery, and Zhou]{wang2022self}
Xuezhi Wang, Jason Wei, Dale Schuurmans, Quoc Le, Ed~Chi, Sharan Narang, Aakanksha Chowdhery, and Denny Zhou.
\newblock Self-consistency improves chain of thought reasoning in language models.
\newblock \emph{arXiv preprint arXiv:2203.11171}, 2022.

\bibitem[Wei et~al.(2022)Wei, Wang, Schuurmans, Bosma, Xia, Chi, Le, Zhou, et~al.]{wei2022chain}
Jason Wei, Xuezhi Wang, Dale Schuurmans, Maarten Bosma, Fei Xia, Ed~Chi, Quoc~V Le, Denny Zhou, et~al.
\newblock Chain-of-thought prompting elicits reasoning in large language models.
\newblock \emph{Advances in neural information processing systems}, 35:\penalty0 24824--24837, 2022.

\bibitem[Welleck et~al.(2019)Welleck, Kulikov, Roller, Dinan, Cho, and Weston]{welleck2019neural}
Sean Welleck, Ilia Kulikov, Stephen Roller, Emily Dinan, Kyunghyun Cho, and Jason Weston.
\newblock Neural text generation with unlikelihood training.
\newblock \emph{arXiv preprint arXiv:1908.04319}, 2019.

\bibitem[Xie et~al.(2024)Xie, Chen, Lee, Mitchell, and Finn]{xie2024calibrating}
Johnathan Xie, Annie~S Chen, Yoonho Lee, Eric Mitchell, and Chelsea Finn.
\newblock Calibrating language models with adaptive temperature scaling.
\newblock \emph{arXiv preprint arXiv:2409.19817}, 2024.

\bibitem[Xu et~al.(2023)Xu, Lee, Sukhbaatar, and Weston]{xu2023some}
Jing Xu, Andrew Lee, Sainbayar Sukhbaatar, and Jason Weston.
\newblock Some things are more cringe than others: Preference optimization with the pairwise cringe loss.
\newblock \emph{arXiv preprint arXiv:2312.16682}, 2023.

\bibitem[Zhang et~al.(2020)Zhang, Duckworth, Ippolito, and Neelakantan]{zhang2020trading}
Hugh Zhang, Daniel Duckworth, Daphne Ippolito, and Arvind Neelakantan.
\newblock Trading off diversity and quality in natural language generation.
\newblock \emph{arXiv preprint arXiv:2004.10450}, 2020.

\bibitem[Zhang et~al.(2024)Zhang, Bao, and Huang]{zhang2024edt}
Shimao Zhang, Yu~Bao, and Shujian Huang.
\newblock Edt: Improving large language models' generation by entropy-based dynamic temperature sampling.
\newblock \emph{arXiv preprint arXiv:2403.14541}, 2024.

\bibitem[Zhu et~al.(2024)Zhu, Hao, He, Ai, and Wang]{zhu2024improving}
Wenhong Zhu, Hongkun Hao, Zhiwei He, Yiming Ai, and Rui Wang.
\newblock Improving open-ended text generation via adaptive decoding.
\newblock \emph{arXiv preprint arXiv:2402.18223}, 2024.

\bibitem[Zhu et~al.(2023)Zhu, Li, Li, Zhao, Li, Jin, and Mei]{zhu2023improving}
Yuqi Zhu, Jia Li, Ge~Li, YunFei Zhao, Jia Li, Zhi Jin, and Hong Mei.
\newblock Improving code generation by dynamic temperature sampling.
\newblock \emph{arXiv e-prints}, pp.\  arXiv--2309, 2023.

\bibitem[Ziegler et~al.(2019)Ziegler, Stiennon, Wu, Brown, Radford, Amodei, Christiano, and Irving]{ziegler2019fine}
Daniel~M Ziegler, Nisan Stiennon, Jeffrey Wu, Tom~B Brown, Alec Radford, Dario Amodei, Paul Christiano, and Geoffrey Irving.
\newblock Fine-tuning language models from human preferences.
\newblock \emph{arXiv preprint arXiv:1909.08593}, 2019.

\end{thebibliography}
\bibliographystyle{icml2022}

\clearpage
\appendix
\section{Task Details}
\label{sec:ultramathstories_details}

\subsection{N-gram Repeat}
We use the Wikitext-2 benchmark.
We use a 50 tokens prefix as the prompt, allowing the LLM to continue generating.
After generating $N=10$ completions per prompt, we rank these completions by 3-gram-repeats.
We then constructed \method{} preference pairs where the sequences with the lowest and highest 3-gram-repeats are selected as the `chosen' and `rejected' sequences respectively. 
We then use \method{} to train the \ADtok{} model.

\subsection{Math (GSM8K)}
\label{app:gsm8k}
For this task, we use the GSM8K math reasoning dataset \citep{cobbe2021training}. We use chain-of-thought prompting \citep{wei2022chain}, where the model is instructed to explain its reasoning before writing a final answer. We train on a random 1,000 sample subset of the full 7,473 samples. We evaluate on the full 1,319 test samples.

The \method{} preference pairs for this dataset are constructed by generating $N=16$ response samples per prompt, where each generation samples a temperature from the original \AD{} distribution (roughly uniform), and then selecting a chosen and rejected sample based on the oracle GSM8K training labels. 

We evaluate the performance of \ADseq{} compared to 6 different fixed temperature decodings: $\tau=\{0.0, 0.2, 0.4, 0.6, 0.8, 1.0\}$. We measure the winrate of each test sample using the ground truth labels from the GSM8K test set. The winrate is computed by comparing the correctness of each method's response. If one method gets it correct and the other does not, the correct method gets awarded 1 point. If both methods generated a correct or incorrect response, then each method gets 0.5 points.

\subsection{General Instruction Following (UltraFeedback)}
\label{app:ultrafeedback}
The full UltraFeedback dataset contains 64k samples. We train on a random subset of 2,000 samples, and test on another random subset of 1,000 samples.

The training preference pairs for this dataset are constructed by generating $N=16$ samples per prompt, where each generation samples a temperature from the original \AD{} distribution (roughly uniform), and selecting a chosen and rejected sample using the best and worst Armo reward model (ArmoRM) \citep{wang2024interpretable} scores, respectively.

We measure the winrate of \ADseq{} generations compared to each the 6 fixed temperature ($\tau$=\{0.0, 0.2, 0.4, 0.6, 0.8, 1.0\}) generations using ArmoRM scores.

\subsection{Creative Writing (Stories)} \label{app:stories}

For this task, we consider a simple creative writing task where the model is prompted to write a short story of a given title. Each prompt in this dataset has the following structure: ``Write a short 200 word story with the following title.\textbackslash n\textbackslash nTitle:[\textsc{TITLE}]''. 
We call this the ``Stories'' task. Each of the 1,000 training and test titles were generated with Llama3.0-70B.

We use the same method as UltraFeedback for constructing training preference pairs and evaluating.

\subsection{Constrained Creative Writing (ConstrainedStories)}
\label{app:constrainedstories}
Each sample has the following structure: ``Write a creative and coherent story with the following title. You must begin each sentence with a word that starts with ``Ab''.\textbackslash n\textbackslash nTitle: [\textsc{TITLE}]''.

The preference pairs are generated as follows. For each prompt, we first generate $N=16$ response samples. 
To select the chosen response, we consider the top 4 ArmoRM scored responses, and then take the one of those that satisfy the constraint the best (has the highest percentage of sentences that start with ``Ab''). Similarly, for the rejected response, we consider the bottom 4 ArmoRM scored responses and take the one of those that least satisfies the constraint.

Winrates are computed in the following way. If a response satisfies the constraint better (i.e., a higher percentage of ``Ab'' start sentences), then it wins. If there is a tie and both responses have the same constraint satisfaction rate, then it is decided by whichever response has a higher ArmoRM score, where the Armo reward model is run using the prompt without the constraint (i.e. ``Write a creative and coherent story with the following title.\textbackslash n\textbackslash nTitle: [\textsc{TITLE}]'').

\section{\ADseq{} Winrate Values}
\label{sec:ad_seq_winrates}

Tables \ref{tab:ultrafeedback_ad_seq}, \ref{tab:stories_ad_seq} and, \ref{tab:gsm8k_ad_seq} show \ADseq{} winrate values on each of the 3 UltraMathStories subtasks.

\begin{table*}[h!]
\centering
\begin{tabular}{l|c|c}
\toprule
Fixed Temp & \makecell{\ADseq{}\\Winrate} & \makecell{Fixed Temp\\Winrate} \\
\midrule
$\tau=0.0$ & 53.10 & 46.90 \\
$\tau=0.2$ & 53.35 & 46.65 \\
$\tau=0.4$ & 50.80 & 49.20 \\
$\tau=0.6$ & 52.15 & 47.85 \\
$\tau=0.8$ & 52.78 & 47.22 \\
$\tau=1.0$ & 54.89 & 45.11 \\
\bottomrule
\end{tabular}
\caption{\textbf{\ADseq{} vs Fixed Temperatures Winrates on the \textit{UltraFeedback} Task.}}
\label{tab:ultrafeedback_ad_seq}
\end{table*}

\begin{table*}[h!]
\centering
\begin{tabular}{l|c|c}
\toprule
Fixed Temp & \makecell{\ADseq{}\\Winrate} & \makecell{Fixed Temp\\Winrate} \\
\midrule
$\tau=0.0$ & 58.75 & 41.25 \\
$\tau=0.2$ & 57.25 & 42.75 \\
$\tau=0.4$ & 57.05 & 42.95 \\
$\tau=0.6$ & 56.65 & 43.35 \\
$\tau=0.8$ & 54.55 & 45.45 \\
$\tau=1.0$ & 52.10 & 47.90 \\
\bottomrule
\end{tabular}
\caption{\textbf{\ADseq{} vs Fixed Temperatures Winrates on the Stories Task.}}
\label{tab:stories_ad_seq}
\end{table*}

\begin{table*}[h]
\centering
\begin{tabular}{l|c|c}
\toprule
Fixed Temp & \makecell{\ADseq{}\\Winrate} & \makecell{Fixed Temp\\Winrate} \\
\midrule
$\tau=0.0$ & 50.68 & 49.32 \\
$\tau=0.2$ & 51.10 & 48.90 \\
$\tau=0.4$ & 51.14 & 48.86 \\
$\tau=0.6$ & 51.40 & 48.60 \\
$\tau=0.8$ & 51.42 & 48.58 \\
$\tau=1.0$ & 51.82 & 48.18 \\
\bottomrule
\end{tabular}
\caption{\textbf{\ADseq{} vs Fixed Temperatures Winrates on the \textit{GSM8K} Task.}}
\label{tab:gsm8k_ad_seq}
\end{table*}

\section{\ADtok{} Winrate Values}
\label{sec:ad_tok_winrates}

Tables \ref{tab:ultrafeedback_ad_tok}, \ref{tab:stories_ad_tok} and, \ref{tab:gsm8k_ad_tok} show \ADseq{} winrate values on each of the 3 UltraMathStories subtasks.

\begin{table*}[h!]
\centering
\begin{tabular}{l|c|c}
\toprule
Fixed Temp & \makecell{\ADtok{}\\Winrate} & \makecell{Fixed Temp\\Winrate} \\
\midrule
$\tau=0.0$ & 49.60 & 50.40 \\
$\tau=0.2$ & 50.70 & 49.30 \\
$\tau=0.4$ & 48.75 & 51.25 \\
$\tau=0.6$ & 49.60 & 50.40 \\
$\tau=0.8$ & 49.25 & 50.75 \\
$\tau=1.0$ & 52.75 & 47.25 \\
\bottomrule
\end{tabular}
\caption{\textbf{\ADtok{} vs Fixed Temperatures Winrates on the \textit{UltraFeedback} Task.}}
\label{tab:ultrafeedback_ad_tok}
\end{table*}

\begin{table*}[h!]
\centering
\begin{tabular}{l|c|c}
\toprule
Fixed Temp & \makecell{\ADtok{}\\Winrate} & \makecell{Fixed Temp\\Winrate} \\
\midrule
$\tau=0.0$ & 54.40 & 45.60 \\
$\tau=0.2$ & 53.40 & 46.60 \\
$\tau=0.4$ & 54.20 & 45.80 \\
$\tau=0.6$ & 52.30 & 47.70 \\
$\tau=0.8$ & 51.10 & 48.90 \\
$\tau=1.0$ & 47.25 & 52.75 \\
\bottomrule
\end{tabular}
\caption{\textbf{\ADtok{} vs Fixed Temperatures Winrates on the \textit{Stories} Task.}}
\label{tab:stories_ad_tok}
\end{table*}

\begin{table*}[h!]
\centering
\begin{tabular}{l|c|c}
\toprule
Fixed Temp & \makecell{\ADtok{}\\Winrate} & \makecell{Fixed Temp\\Winrate} \\
\midrule
$\tau=0.0$ & 49.66 & 50.34 \\
$\tau=0.2$ & 50.08 & 49.92 \\
$\tau=0.4$ & 50.11 & 49.89 \\
$\tau=0.6$ & 50.38 & 49.62 \\
$\tau=0.8$ & 50.49 & 49.51 \\
$\tau=1.0$ & 51.55 & 48.45 \\
\bottomrule
\end{tabular}
\caption{\textbf{\ADtok{} vs Fixed Temperatures Winrates on the \textit{GSM8K} Task.}}
\label{tab:gsm8k_ad_tok}
\end{table*}

\begin{table*}[]
\centering
\begin{tabular}{c|c|c|c}
\toprule
Fixed Temp & \makecell{\ADtok{} \\Constraint Winrate} & \makecell{\ADtok{}\\ArmoRM Winrate} & \makecell{\ADtok{}\\Avg Winrate} \\
\midrule
$\tau=0.0$ & 50.95 & 52.55 & 51.75 \\
$\tau=0.2$ & 53.70 & 49.50 & 51.60 \\
$\tau=0.4$ & 58.05 & 48.25 & 53.15 \\
$\tau=0.6$ & 68.05 & 41.05 & 54.55 \\
$\tau=0.8$ & 77.85 & 36.45 & 57.15 \\
$\tau=1.0$ & 87.80 & 31.50 & 59.65 \\
\bottomrule
\end{tabular}
\caption{\textbf{\ADtok{} Constrained Creative Writing Individual Winrates.} Here we show the individual winrates of the \ADtok{} for both constraint following and ArmoRM score. The \ADtok{} learns to follow the constraint better than all fixed temperatures, but as we compare to higher fixed temperatures, the story winrate goes down because it follows the constraint better.}
\label{tab:constrained_full_winrates}
\end{table*}

\subsection{Constrained Creative Story Writing Example Temperatures}
\begin{figure*}
\centering
\begin{subfigure}[b]{0.40\textwidth}
   \includegraphics[width=1\linewidth]{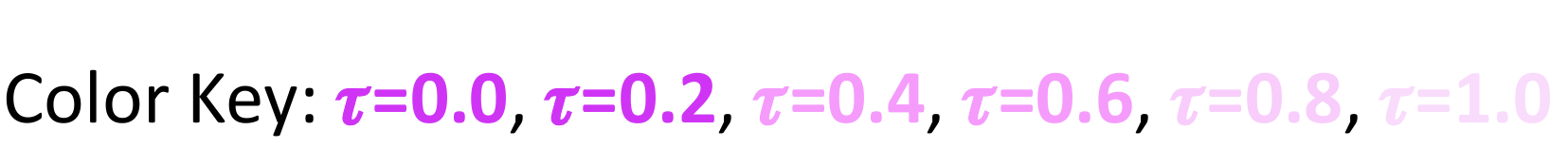}
\end{subfigure}

\begin{subfigure}[b]{1.0\textwidth}
   \fbox{\includegraphics[width=1\linewidth]{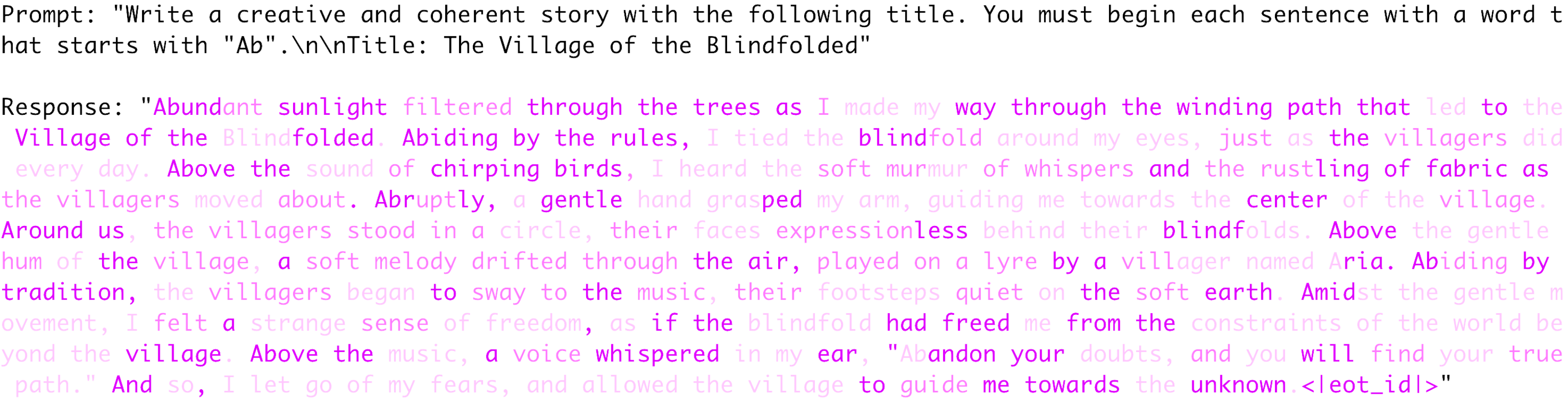}}
\end{subfigure}

\caption{\textbf{\ADtok{} predicted temperatures for Constrained Creative Story Writing.} We demonstrate an example of \ADtok{} predicted temperatures ($\tau$) on the constrained creative story writing task for the prompt \textit{``Write a creative and coherent story with the following title. You must begin each sentence with a word that starts with ``Ab''.\textbackslash n\textbackslash nTitle: The Village of the Blindfolded''}. We can see that the model is more greedy ($\tau$ close to 0.0) when generating the constraint tokens (All sentences must begin with words that start with ``Ab''), and less greedy ($\tau$ close to 1.0) on all other tokens.}
\label{fig:constrained_story_ex}
\end{figure*}

\autoref{fig:constrained_story_ex} shows an example of the predicted temperature values for the \ADtok{} model trained on constrained creative story generation.

\begin{figure*}[t]
  \centering
  \begin{subfigure}[b]{0.33\textwidth}
    \includegraphics[width=\textwidth]{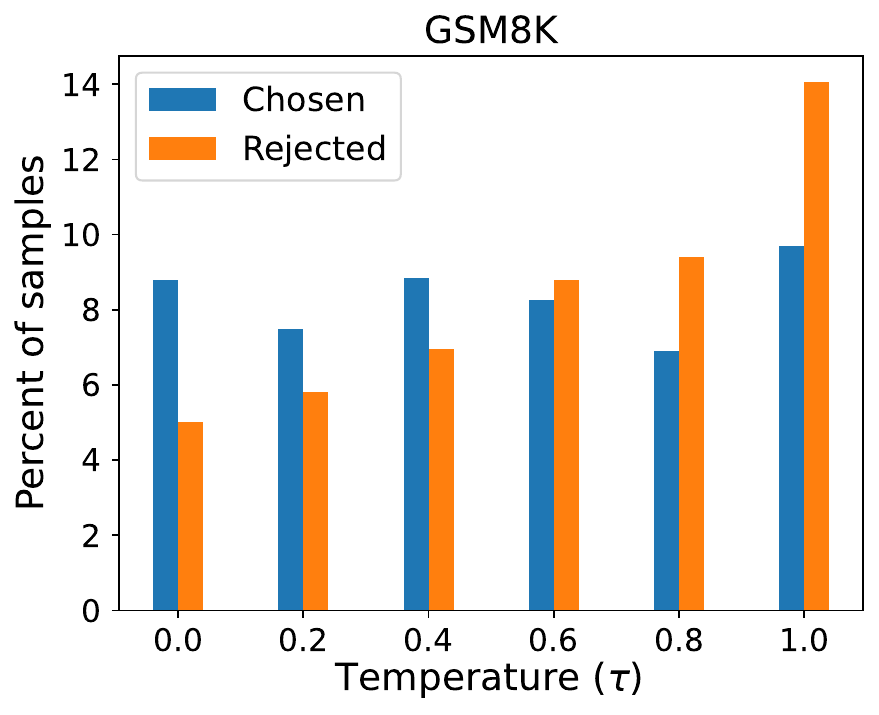}
  \end{subfigure}
  \begin{subfigure}[b]{0.33\textwidth}
    \includegraphics[width=\textwidth]{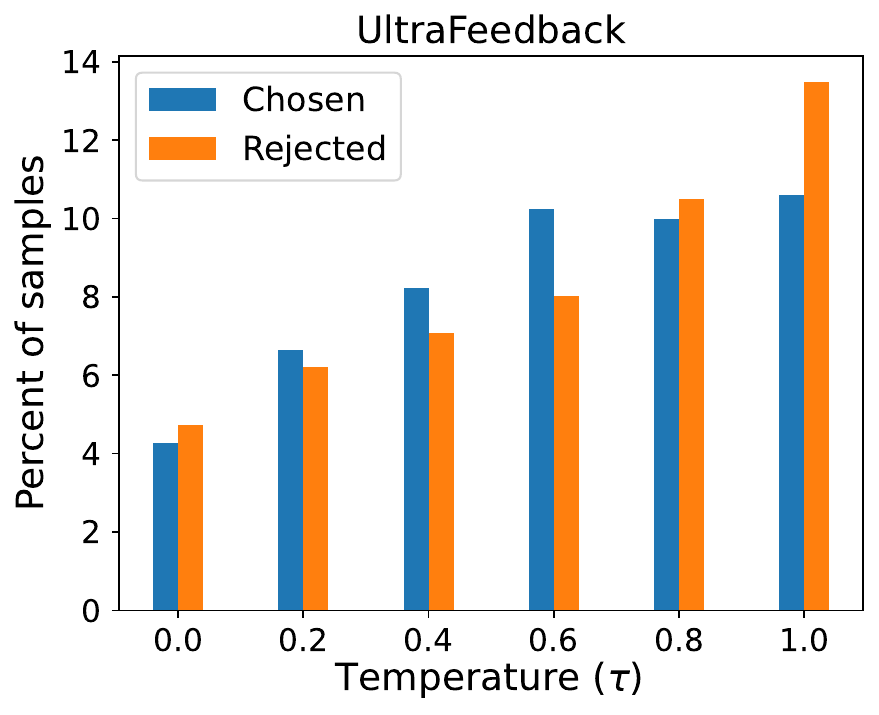}
  \end{subfigure}
  \begin{subfigure}[b]{0.33\textwidth}
    \includegraphics[width=\textwidth]{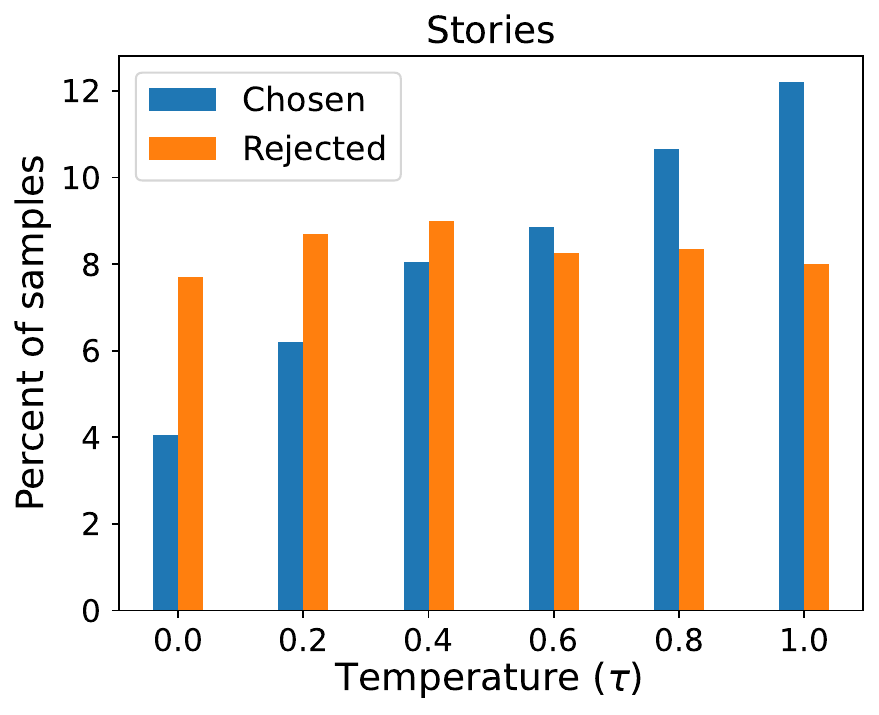}
  \end{subfigure}
  \caption{\textbf{\ADseq{} Training Preference  Distributions}. Here we show the percentage of samples in the training set that are chosen or rejected for each of the 6 different temperateure ($\tau$) values. The \method{} loss uses both chosen and rejected responses, and the ratio of chosen to rejected is an important factor for learning the right temperature. A vanilla negative log-likelihood loss only uses the chosen responses, which leads to suboptimal temperature predictions since high temperature values are the most chosen regardless of the task.}
  \label{fig:training_distribitions}
\end{figure*}

\begin{table*}[h!]
\centering
\begin{tabular}{p{16cm}}
\multicolumn{1}{c}{{\color{my_purple} Predicted $\tau=0.0$}}\\
\toprule
In this task, given a sentence in the English language, your task is to convert it into the Thai language.\\Problem:The secondary principals' association head, Graham Young, said: \"The NCEA system put pressure on schools to accumulate credits - and the easiest way to do that was to encourage students into internally assessed unit standards.\\Solution: \\
\hline
You are given a math word problem and you are supposed to apply multiple mathematical operators like addition, subtraction, multiplication, or division on the numbers embedded in the text to answer the following question and then only report the final numerical answer.\\\\Input: Consider Input: debby makes 67 pancakes . she adds blueberries to 20 of them and bananas to 24 of them . the rest are plain . how many plain pancakes are there ?\\
\hline
You have been tasked with arranging a group of travelers, each with different preferences and needs, onto various modes of transportation. There are four modes of transportation available: A, B, C, and D. Each mode has its own unique features and limitations. The travelers and their preferences are as follows:\\1. Alice: Is afraid of flying and prefers to take mode C or D\\2. Bob: Can only travel by mode A due to motion sickness\\3. Charlie: Wants to take mode B because it has the shortest travel time\\4. Dave: Needs to take mode D because he has a lot of luggage\\5. Ellie: Wants to take mode A because she enjoys the scenic route\\Your task is to assign each traveler to the mode of transportation that best suits their needs and preferences. Keep in mind that each mode of transportation can only accommodate a certain number of people, and some modes may have already reached their capacity. Can you solve this puzzle and successfully group the travelers onto their preferred modes of transportation?" \\
\bottomrule
\\
\\
\multicolumn{1}{c}{{\color{my_purple}Predicted $\tau=1.0$}}\\
\toprule

Write a 70,000 word fantasy novel about a hidden world of magic and mythical creatures. The main character must be a human who discovers this world and becomes involved in a conflict between the magical creatures. The novel should have a fast-paced plot with plenty of action and suspense. The style should be descriptive and immersive, with detailed descriptions of the magical world and its inhabitants. The novel should also explore themes such as the nature of power and the importance of loyalty and friendship.\\

\hline
Write me a 1000 word ghost story in a campfire setting \\

\hline
Write a story about Ego Must, a prominent innovator with technology who leverages his vast wealth to communicate his views. However, despite being exceptionally smart he seems to not understand the basics when it comes to the 'us and them' problem that is at the root of a lot of human conflict. \\
\bottomrule

\end{tabular}
\caption{Examples of \textbf{\ADseq{} Predicted Temperatures ($\tau$) on UltraFeedback.}
Here we show examples of UltraFeedback test prompts where the \ADseq{} model predicted $\tau \in \{0.0, 1.0\}$. We can see that the $\tau=0.0$ prompts require factual, deterministic responses, and the $\tau=1.0$ prompts require creative, stochastic responses. This shows generalization outside of the GSM8K and Stories subtasks to specific prompts within UltraFeedback.
}
\label{tab:ultrafeedback_examples_appendix}
\end{table*}

\end{document}